# Locating Tables in Scanned Documents for Reconstructing and Republishing

MAC Akmal Jahan
Post Graduate Institute of Science
University of Peradeniya
Peradeniya, Sri Lanka

Roshan G Ragel
Department of Computer Engineering
University of Peradeniya
Peradeniya, Sri Lanka

***Abstract* — Pool of knowledge available to the mankind depends on the source of learning resources, which can vary from ancient printed documents to present electronic material. The rapid conversion of material available in traditional libraries to digital form needs a significant amount of work if we are to maintain the format and the look of the electronic documents as same as their printed counterparts. Most of the printed documents contain not only characters and its formatting but also some associated non text objects such as tables, charts and graphical objects. It is challenging to detect them and to concentrate on the format preservation of the contents while reproducing them. To address this issue, we propose an algorithm using local thresholds for word space and line height to locate and extract all categories of tables from scanned document images. From the experiments performed on 298 documents, we conclude that our algorithm has an overall accuracy of about 75% in detecting tables from the scanned document images. Since the algorithm does not completely depend on rule lines, it can detect all categories of tables in a range of scanned documents with different font types, styles and sizes to extract their formatting features. Moreover, the algorithm can be applied to locate tables in multi column layouts with small modification in layout analysis. Treating tables with their existing formatting features will tremendously help the reproducing of printed documents for reprinting and updating purposes.**

## I. Introduction

A pool of knowledge base depends on the source of learning resources, which can vary from ancient printed documents to present electronic material. Traditional libraries play a significant role in the dissemination and safekeeping of such material. The rapid conversion of material available in traditional libraries to digital form needs a huge amount of manual effort, if we are to maintain the format and the look of the electronic documents as same as their printed counterparts. Therefore, proper digitization of printed documents plays a significant role in building digital libraries. Modifying, re-printing and searching are some important functions beyond mere digitization to maintain the digital library continuously. Therefore, we can conclude that the requirement of digitization here is converting printed documents into editable digital format, while preserving both the content and the format of the documents.

In recent past, the OCR technology worked well in the digitization process of scripts in different languages.

Typically, most of the OCR approaches focus only on character recognition of the script and it advances to focus on extraction of format features such as font size, plain text, italics, bold with different font sizes. Typical process of converting printed documents into an editable format is performed by scanning documents to images and converting them to editable text by using OCR techniques. Layout analysis, character recognition and language modelling are the important processes in OCR for a particular script. However, most of the printed documents contain not only characters but also some associated non text objects such as tables, charts and graphics in an image format. Most of the existing OCR techniques face challenges in detecting these kinds of objects during the digitization process of printed documents. Particularly presence of such objects affects the layout analysis, which is the initial step of the OCR process. For example, if the presence of a table in a document is not identified by OCR, there will be problems in identifying lines and figures inside the table and they will be treated as regular sentences. Practically, this would produce more erroneous results in text analysis in any type of script. In addition to the recognition process of the contents in the documents, editing process needs to identify the formats of the contents to maintain the uniformity within pre-processing and post processing such as updating and reproducing the contents. Most of the OCR techniques do not concentrate on the format preservation of the content. Therefore, this will be a challenging task when reproducing printed documents for reprinting or republishing purpose.

Therefore, we need to identify a way to solve the problem of preserving and reproducing documents with features discussed above. In this paper, we focus on locating and extracting different types of tables with text portion in documents. Treating tables with their existing formatting features will do a great job when reproducing the documents for future need.

The rest of the paper is organized as follows. In Section II, we present a literature survey on table detection and recognition and in Section III, we present our methodology. In Section IV, we present our experimental results and a discussion on the results in Section V and we conclude the paper in Section VI.

## II. Literature Survey

Generally, tables vary in structure from regular text. They can be designed by using ruled line, off-line, decorated line or without using lines. However, the basic structure of a table does not depend on the line but in its building blocks such as rows, columns and fields. Tables have physical and logical structures [1]. The physical structure determines the regions of a table in a document whereas the logical structure determines the relational information of the table such as constituent parts and how they form a table. Therefore, all parts in a table have both physical and logical structures, which are used in table region detection and table structure recognition respectively. Most of the recent research on table recognition assumes that the table region is already known and the work focused only on extraction of its logical structure. Since the existing table recognition systems lacks the facility to locate tables and the other non-text objects, it affects the layout analysis of the document image. To overcome the problem, tables can be handled in two steps: a) table detection (or locating tables), and b) table recognition.

Table detection and recognition in documents can be handled by two different approaches based on the input document type such as scanned image and electronic text documents such as pdf, HTML, word and ASCII [5-7]. Recently, there are many researches focus on table recognition rather than locating the table (or table detection) [2-3], [5-11]. However, there are some work focus on table detection and not table recognition. Rarely any work cover both table detection and recognition [10, 17]. In the table detection process, there are several ways that use geometric features such as ruled lines [12,13,15,19], pixel distributions, white gaps [4], header and trailer pattern of table [21]. Most of the past works only focus on single column page layout, whereas some recent work focuses on multi column page layout [8], [16].

Namboodiri addressed the problem of table detection and recognition for online hand written documents [17] whereas Jin Chen and Daniel addressed this issue for off-line handwritten documents [18]. Laurintine and Vaida use horizontal and vertical ruling lines to identify tables and exclude non-tabular areas from the printed documents [10]. Hu et al. proposed a method, which does not depend on ruling lines and medium independent [15]. Gatos et al. [13] locate and reconstruct tables with intersection pairs, which are detected by horizontal and vertical lines' intersection points. Tables are reconstructed by drawing the lines. Mandal et al. [4] proposed an algorithm, which assumes the presence of substantially larger gaps between columns and table fields. This system can locate the tables if they do not contain any ruling lines or which should be removed in pre-processing. However, this algorithm does not show any way to handle tables when two or more of them appear side by side. Also tables with multi line headings or heterogeneous placement of cells can result in erroneous detection. In contrast to the above methods, our approach can locate tables with their existing formats and can resolve the problem in multiline heading as well.

Harit and Bansal [21] proposed a method for table detection using header and trailer patterns. However, the pattern depends on cognitive cues rather than functional meaning. Shafait et al. [16] and Smith et al. [22] have used tab-stop detection for doing the layout analysis of document images and then used the alignment information of columns for locating the tables. This algorithm does not work when full page tables are present. Since we are focussing on all type of tables in a page and not concern about column layout analysis of the page, our algorithm can handle this type of tables easily.

Most of the existing methods focus only on table detection through several assumptions such as i) there is a specific layout such as single column or multi column and white background in document image; ii) table is constructed with horizontal and vertical ruled-line; and iii) there are no decorated lines or complex table structure.

From the overall analysis, all past work handle only specific page layout and specific table category rather than all range of tables in different categories. Therefore, we need to identify a novel way to locate all types of tables from text in scanned documents. The algorithms proposed in this paper can address detection of all categories of tables with or without ruled lines.

## III. METHODOLOGY

The sample document images consist of normal text lines, headings, a wide range of tables, equations, header, footer and page numbers with different font sizes and styles. The background is white and grey coloured. The assumptions we made here are as follows:

- There are no graphics and watermarks in the document images.
- The minimum height of a table is equal to or greater than 3 regular text lines. With this assumption, most of the heading and equation parts can be eliminated from being detected as tables.
- In a standard full lengthy text line, there would be more number of characters compared to other text lines.
- The document image is single column layout and for a professional document, there is no text or image portion on left or right side of a table (that is, tables are in line with text).

The methodology as presented in Figure 1 follows a number of stages and they are: 1) data collection and pre-processing of the scanned documents; 2) threshold computation for word space and line height; and 3) extraction of the table with and without rule lines. Let us look at details of each stage in the rest of this section.

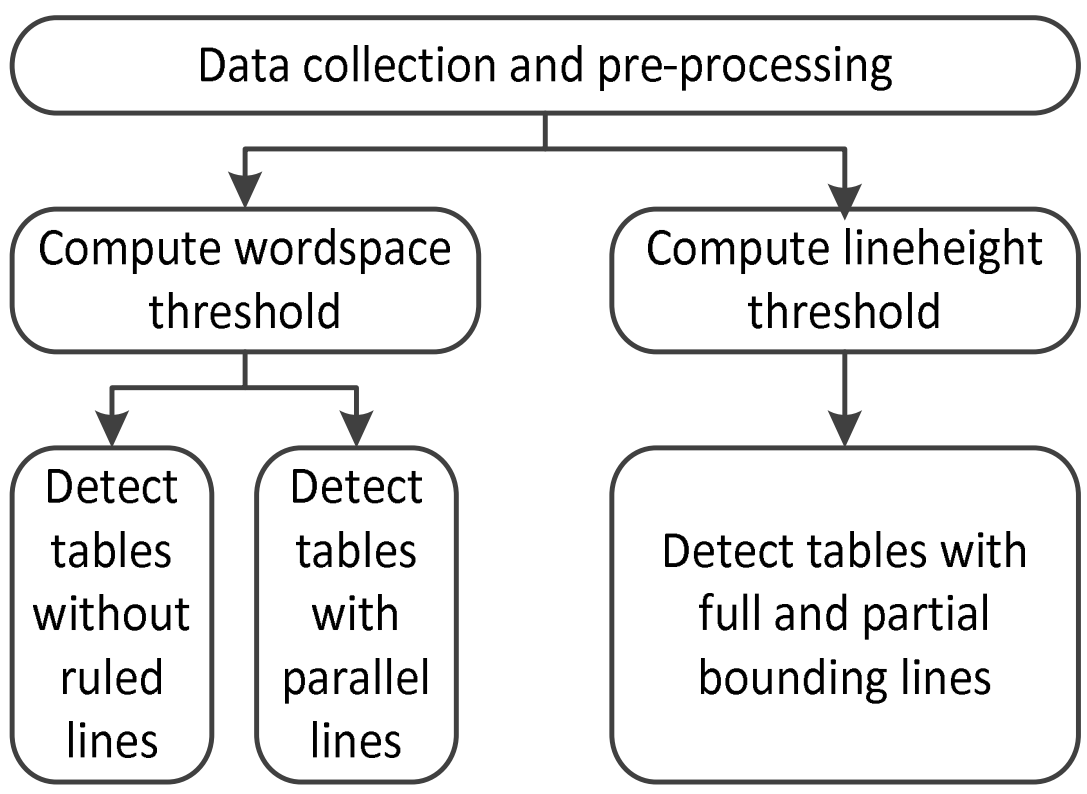

Fig. 1: System modules

### A. *Our Algorithm*

The algorithm is driven from several steps of computations. Initially, all horizontal lines in the image were extracted using X projection profile and their heights were calculated using Y projection profile. Spaces between characters and words were computed in all lines and the line which had the most number of gaps was considered as a standard text line. The assumption we made here is if there are more number of gaps, there would be more number of characters and that would be a standard text line.

Word space is assigned in all text line by considering the maximum size of the space in each line. The threshold value for a word space in a standard text line is computed as in equation (1).

$$WS < ws < 2*WS \quad (1)$$

*WS*- word space
*ws*-threshold for word space

Similarly, line height (*LH*) is assigned for all lines. The threshold value for the standard line height is computed as in equation (2).

$$LH < lh < 1.5*LH \quad (2)$$

*LH*-line height
*lh*-threshold for line height

The algorithm for the table detection is given as follows:

```
for x=1 to N do
   if LH>=3*lh AND number of gaps= 0
        locate table type A
   end if
   if WS>ws AND LH<=lh
        locate table type B and C
   end if
   if number of gaps=0 AND LH<lh AND
   AND WS(x-1)> ws OR WS(x+1)>ws
        locate parallel rule line
   end if
end for
```

Where

N- number of lines
WS- word space
LH- line height
lh- threshold for line height
ws- threshold for word space
x- line number

In the given algorithm, the height of each line is compared with the threshold of the line height. Here we make an assumption that the minimum height of a table might be at least 3 times greater than the normal text line and there is no space in the line. Therefore, the tables with all bounding box could act as a single line without gaps and could be located.

Similarly, the maximum size of the inter word space is compared with the threshold of word space to detect columns portion in a table. From this step the tables without rule line and parallel lines could be located. To detect the rule line corresponding to the table, the gaps in a line is assumed to be zero and line height is less than the line height threshold. Mean time this line is checked with previous and next text line whether those lines are in table portion.

### B. *Experimental Procedure*

The document pages were scanned from several documents such as books, research papers and journals into an image format such as tiff, jpg or png. Initial part of this work is to extract different categories of tables from different document images. Around 300 scanned images were considered for the testing process. The scanned documents with tables were categorized into three types: A) tables with fully and partially bounding lines; B) tables with parallel lines; and C) tables without rule line. Pre-processing of a scanned document plays a significant role in the detection process. It involves binarization, noisy border removal, and enhancement. Binarization in which colour images or grey-scale images are converted to binary image using adaptive threshold technique [23]. Dilation method is applied for the enhancement with the structuring element. The pre-processed images are applied with the algorithm to locate tables.

Table 1.
**Health Status by Selected Characteristics: 2010**
(Numbers in thousands. Only people in the noninstitutionalized population)

| Characteristic | Total number | Health status (percent) | | | | | | | | | |
|---|---|---|---|---|---|---|---|---|---|---|---|
| | | Excellent | | Very Good | | Good | | Fair | | Poor | |
| | | Esti-mate | Stan-dard error[1] | Esti-mate | Stan-dard error[1] | Esti-mate | Stan-dard error[1] | Esti-mate | Stan-dard error[1] | Esti-mate | Stan-dard error[1] |
| **All people** | **304,814** | **32.7** | **0.32** | **32.9** | **0.30** | **24.1** | **0.24** | **7.9** | **0.10** | **2.4** | **0.06** |
| **Sex** | | | | | | | | | | | |
| Male | 149,421 | 33.9 | 0.35 | 33.1 | 0.35 | 23.4 | 0.31 | 7.4 | 0.14 | 2.2 | 0.09 |
| Female | 155,393 | 31.6 | 0.35 | 32.7 | 0.33 | 24.7 | 0.26 | 8.4 | 0.15 | 2.5 | 0.09 |
| **Race and Hispanic origin[2]** | | | | | | | | | | | |
| White, non-Hispanic | 197,798 | 32.7 | 0.35 | 33.4 | 0.34 | 23.6 | 0.29 | 7.9 | 0.13 | 2.4 | 0.08 |
| Black, non-Hispanic | 36,154 | 29.8 | 0.67 | 32.1 | 0.62 | 25.3 | 0.66 | 10.1 | 0.34 | 2.7 | 0.18 |
| Other, non-Hispanic | 21,231 | 35.3 | 0.86 | 31.6 | 0.79 | 23.4 | 0.68 | 7.7 | 0.35 | 2.1 | 0.16 |
| Hispanic | 49,631 | 33.8 | 0.76 | 32.3 | 0.69 | 25.4 | 0.60 | 6.5 | 0.25 | 1.9 | 0.15 |
| **Age** | | | | | | | | | | | |
| Under 18 years | 74,802 | 59.4 | 0.62 | 27.3 | 0.56 | 11.7 | 0.34 | 1.4 | 0.09 | 0.3 | 0.05 |
| 18 years and older | 230,012 | 24.0 | 0.29 | 34.8 | 0.28 | 28.1 | 0.28 | 10.0 | 0.13 | 3.0 | 0.08 |
| Less than 65 years | 265,947 | 36.2 | 0.36 | 33.9 | 0.33 | 22.1 | 0.26 | 6.1 | 0.11 | 1.7 | 0.06 |
| 18 to 24 years | 29,543 | 42.0 | 0.74 | 35.6 | 0.62 | 18.2 | 0.55 | 3.6 | 0.27 | 0.5 | 0.10 |
| 25 to 44 years | 81,068 | 30.6 | 0.47 | 38.6 | 0.44 | 24.0 | 0.40 | 5.5 | 0.19 | 1.3 | 0.09 |
| 45 to 64 years | 80,534 | 18.1 | 0.36 | 34.8 | 0.42 | 31.4 | 0.41 | 11.9 | 0.25 | 3.8 | 0.15 |
| 65 years and over | 38,867 | 9.2 | 0.29 | 26.3 | 0.49 | 37.4 | 0.50 | 20.3 | 0.43 | 6.8 | 0.24 |
| **Family income as a percentage of poverty threshold[3]** | | | | | | | | | | | |
| Less than 200 percent | 111,544 | 29.7 | 0.45 | 29.8 | 0.44 | 26.3 | 0.42 | 10.5 | 0.19 | 3.7 | 0.13 |
| Less than 100 percent (in poverty) | 49,508 | 31.5 | 0.63 | 29.8 | 0.59 | 25.2 | 0.56 | 9.8 | 0.29 | 3.7 | 0.19 |
| 100 percent to less than 200 percent | 62,036 | 28.3 | 0.60 | 29.9 | 0.56 | 27.1 | 0.56 | 11.0 | 0.26 | 3.7 | 0.17 |
| 200 percent or higher | 192,670 | 34.4 | 0.36 | 34.8 | 0.34 | 22.8 | 0.26 | 6.4 | 0.11 | 1.6 | 0.06 |
| 200 percent to less than 300 percent | 52,197 | 30.5 | 0.60 | 32.5 | 0.60 | 25.8 | 0.50 | 8.9 | 0.25 | 2.4 | 0.14 |
| 300 percent to less than 400 percent | 41,535 | 32.2 | 0.67 | 34.4 | 0.58 | 24.9 | 0.57 | 6.8 | 0.27 | 1.7 | 0.14 |
| 400 percent or higher | 98,938 | 37.4 | 0.46 | 36.1 | 0.44 | 20.4 | 0.31 | 5.0 | 0.14 | 1.1 | 0.07 |

[1] Standard error estimates were calculated using replicate weights, Fay's Method.

[2] Federal surveys now give respondents the option of reporting more than one race. Therefore, two basic ways of defining a race group are possible. A group, such as Black, may be defined as those who reported Black and no other race (the race-alone or the single-race concept) or as those who reported Black regardless of whether they also reported another race (the race-alone-or-in-combination concept). Hispanics may be any race. The body of this report (text, figures, and tables) shows data for people who reported they were a single race. Use of the single-race concept does not imply that it is the preferred method of presenting or analyzing data. The Census Bureau uses a variety of approaches. In this report, the term "non-Hispanic White" refers to people who are not Hispanic and reported White and no other race. "Non-Hispanic Black" refers to people who are not Hispanic and reported Black and no other race. "Non-Hispanic Other" refers to people who are not Hispanic and reported Asian alone, Pacific Islander alone, American Indian alone, Alaskan Native alone, or multiple races.

[3] The poverty universe is slightly smaller than that reported under "All People" as it excludes people less than 15 years old with no cohabitating relatives, who are not asked income questions.

Source: U.S. Census Bureau, Survey of Income and Program Participation, 2008 Panel, wave 7 topical module and core survey data. For information on confidentiality protection and sampling and nonsampling error, see <www.census.gov/sipp/source.html>.

Fig. 2. Type A-Table with partially bounding lines

**Table 3. Reporting Facilities for African American Women with Breast Cancer, Nebraska, 1999-2008**

| Rank | Facility | County | Number | Percent |
|---|---|---|---|---|
| 1 | Alegent Health | Douglas | 109 | 29% |
| 2 | The Nebraska Medical Center | Douglas | 101 | 27% |
| 3 | Creighton University Medical Center | Douglas | 78 | 21% |
| 4 | Nebraska Methodist Hospital | Douglas | 37 | 10% |
| 5 | Bryan LGH Medical Center | Lancaster | 24 | 6% |
| 6 | Ehrling Bergquist Hospital | Sarpy | 8 | 2% |
| 7 | Saint Elizabeth Regional Medical Center | Lancaster | 8 | 2% |
| 8 | All Other | N/A | 11 | 3% |
| Total | | | 376 | 100% |

Data Source: Nebraska Cancer Registry.

**Staging**

African American women were significantly more likely than Caucasian women to be diagnosed at a later stage (regional or distant stage) (p=0.0002). About 4 in 10 African American women (37.5%) were diagnosed at either a regional or distant stage, compared to 3 in 10 Caucasian women (28.2%) (Table 4).

**Table 4. Stage at Diagnosis among African American and Caucasian Women Diagnosed with Breast Cancer, Nebraska, 1999-2008**

| | African American | | Caucasian | |
|---|---|---|---|---|
| **Stage** | **Number** | **Percent** | **Number** | **Percent** |
| In-Situ | 66 | 17.6 | 2356 | 17.0 |
| Local | 163 | 43.4 | 7123 | 51.3 |
| Regional | 120 | 31.9 | 3459 | 24.9 |
| Distant | 21 | 5.6 | 462 | 3.3 |
| Unstaged | 6 | 1.6 | 477 | 3.4 |
| Total | 376 | 100.1 | 13877 | 99.9 |

Data Source: Nebraska Cancer Registry.

Fig. 3. Type A-Table with fully bounding lines

**Table 1.** Source retrieval results with respect to retrieval performance and cost-effectiveness.

| Team (alphabetical order) | Downloaded Sources | | | Total Workload | | Workload to 1st Detection | | No Detection | Runtime |
|---|---|---|---|---|---|---|---|---|---|
| | $F_1$ | Precision | Recall | Queries | Downloads | Queries | Downloads | | |
| Elizalde | 0.17 | 0.12 | 0.44 | 44.50 | 107.22 | 16.85 | 15.28 | 5 | 241.7 m |
| Gillam | 0.04 | 0.02 | 0.10 | 16.10 | 33.02 | 18.80 | 21.70 | 38 | **15.1 m** |
| Haggag | 0.44 | **0.63** | 0.38 | 32.04 | **5.93** | 8.92 | **1.47** | 9 | 152.7 m |
| Kong | 0.01 | 0.01 | **0.65** | 48.50 | 5691.47 | 2.46 | 285.66 | **3** | 4098.0 m |
| Lee | 0.35 | 0.50 | 0.33 | 44.04 | 11.16 | 7.74 | 1.72 | 15 | 310.5 m |
| Nourian | 0.10 | 0.15 | 0.10 | **4.91** | 13.54 | **2.16** | 5.61 | 27 | 25.3 m |
| Suchomel | 0.06 | 0.04 | 0.23 | 12.38 | 261.95 | 2.44 | 74.79 | 10 | 1637.9 m |
| Veselý | 0.15 | 0.11 | 0.35 | 161.21 | 81.03 | 184.00 | 5.07 | 16 | 655.3 m |
| Williams | **0.47** | 0.55 | 0.50 | 116.40 | 14.05 | 17.59 | 2.45 | 5 | 1163.0 m |

queries without submitting them and submit all queries computed when 20% of the words in the suspicious document are contained in the queries without computing any new query afterwards. The retrieval scores of ChatNoir are used to compute the sum of these scores for a document over all queries and in the end the 15 documents with highest sum are downloaded (however, this is not consistent with the overall workload we measured, which is around 31 downloads per suspicious document). Williams et al. [44] download the top-3 documents whose snippets share at least five word 5-grams with the suspicious document.

Fig. 4. Type B-Table with parallel lines

In Cape Town, there were 1 704 child deaths under 5 years of age in 2004 (Table 2.2). The majority of the deaths occurred in the postnatal period (49.9%) and a sizable proportion in the early neonatal period (22.7%). The age distribution of the deaths in the Boland/Overberg is shown in Table 2.3 and it is very similar to that observed in Cape Town.

**Table 2.2: Age distribution of deaths under 5 years, Cape Town, 2004**

| Age of deaths | Number | % |
|---|---|---|
| 0–6 days | 386 | 22.7 |
| 7–27 days | 154 | 9.0 |
| 1–11 months | 851 | 49.9 |
| 1–4 years | 313 | 18.4 |
| **Total** | **1 704** | **100.0** |

Source: Unpublished data from City of Cape Town

**Table 2.3: Age distribution of deaths under 5 years, Boland/Overberg, 2005**

| Age group | Number | % |
|---|---|---|
| 0–6 days | 167 | 21.9 |
| 7–27 days | 69 | 9.0 |
| 1–11 months | 368 | 48.2 |
| 1–4 years | 159 | 20.8 |
| **Total** | **763** | **100.0** |

Source: Unpublished data from Boland/Overberg

***Health and Demographic Surveillance Sites***

South Africa has three Health and Demographic Surveillance Sites (HDSSs) where

Fig. 5. Type B-Table with parallel lines

Appendix B

**Microsoft SQL Server 2000, SQL Server 2005, and Oracle Database 10*g* Comparison**

The following table shows the statements that are sent to the supported databases when database transactions are handled.

| Microsoft Dynamics AX 4.0 | SQL Server 2000 | SQL Server 2005 | Oracle Database 10*g* |
|---|---|---|---|
| First *ttsbegin* statement | SET TRANSACTION ISOLATION LEVEL READ COMMITTED | SET TRANSACTION ISOLATION LEVEL READ COMMITTED | No statement sent |
| First SQL DML statement inside a transaction scope | SET IMPLICIT_ TRANSACTIONS ON | SET IMPLICIT_ TRANSACTIONS ON | No statement sent |
| Final *ttscommit* statement | COMMIT TRANSACTION | COMMIT TRANSACTION | COMMIT |
| | SET TRANSACTION I SOLATION LEVEL READ UNCOMMITTED | SET TRANSACTION ISOLATION LEVEL READ COMMITTED | |
| *ttsabort* statement | ROLLBACK TRANSACTION | ROLLBACK TRANSACTION | ROLLBACK |
| | SET TRANSACTION ISOLATION LEVEL READ UNCOMMITTED | SET TRANSACTION ISOLATION LEVEL READ COMMITTED | |
| First SQL DML statement outside a transaction scope | SET IMPLICIT_ TRANSACTIONS OFF | SET IMPLICIT_ TRANSACTIONS OFF | No statement sent |
| *selectLocked(false)* | WITH (NOLOCK) hint added to SELECT statement | Not supported, so no hint added | Not supported, so no hint added |
| Select *optimisticlock concurrencyModel (ConcurrencyModel:: OptimisticLock)* | WITH (NOLOCK) hint added to SELECT statement | No hint | No hint |
| Select *pessimisticlock concurrencyModel (ConcurrencyModel:: PessimisticLock)* | WITH (UPDLOCK) hint added to SELECT statement | WITH (UPDLOCK) hint added to SELECT statement | FOR UPDATE OF clause added to SELECT statement |

485

Fig. 6. Type C-Table without rule lines

```
...
Function        man                          param([string]$Name,[stri...
Function        md                           param([string[]]$paths); ...
Function        mkdir                        param([string[]]$paths); ...
Function        more                         param([string[]]$paths); ...
...
Function        prompt                       'PS ' + $(Get-Location) +...
...
```

Excellent. We could do the same for aliases, applications, external scripts, filters, and scripts. Also note that Get-Command allows you search for cmdlets based on either a Noun or a Verb. There's a more compact form that most of the PowerShell regulars use instead of these parameters though:

```
PS> Get-Command write-*

CommandType     Name                         Definition
-----------     ----                         ----------
Cmdlet          Write-Debug                  Write-Debug [-Message] <S...
Cmdlet          Write-Error                  Write-Error [-Message] <S...
Cmdlet          Write-Host                   Write-Host [[-Object] <Ob...
Cmdlet          Write-Output                 Write-Output [-InputObjec...
Cmdlet          Write-Progress               Write-Progress [-Activity...
Cmdlet          Write-Verbose                Write-Verbose [-Message] ...
Cmdlet          Write-Warning                Write-Warning [-Message] ...
```

You can swap the wildcard char to find all verbs associated with a particular noun (usually the more useful search):

```
PS> Get-Command *-object

CommandType     Name                         Definition
-----------     ----                         ----------
Cmdlet          Compare-Object               Compare-Object [-Referenc...
Cmdlet          ForEach-Object               ForEach-Object [-Process]...
Cmdlet          Group-Object                 Group-Object [[-Property]...
Cmdlet          Measure-Object               Measure-Object [[-Propert...
Cmdlet          New-Object                   New-Object [-TypeName] <S...
Cmdlet          Select-Object                Select-Object [[-Property...
Cmdlet          Sort-Object                  Sort-Object [[-Property] ...
Cmdlet          Tee-Object                   Tee-Object [-FilePath] <S...
Cmdlet          Where-Object                 Where-Object [-FilterScri...
```

Finally, we can pass a name to Get-Command to find out if this name will be interpreted as a command and if so, what type of command: alias, application, cmdlet, external script, filter, function or script. In this usage, Get-Command is like the UNIX *which* command on steroids. Let me show you what I mean:

Page 4

Fig. 7. Type C-Table without rule lines

In this section, we are showing some examples for each type of tables we defined. Figures 2 and 3 depict example tables of type A with fully and partial bounding lines and Figures 4 and 5 depict example tables of type B with parallel lines. Figures 6 and 7 depict example tables of type C without rule lines.

## IV. Results and Discussion

The categorized tables presented in the last section are tested with the proposed algorithm.

Table I illustrates that the different type of table detection from tables with fully and partially bounding lines to tables without ruling lines. Category A shows the highest percentage of detection whereas table with parallel line shows lower performance in detection.

From Figure 8, it can be seen that all tables produce accuracy of detection better than 67%. Tables with fully or partially bounding lines shows higher accuracy with 82.7% of detection compared to the others. These type of tables can be extracted easily as it is bounded with at least single vertical line and we compare their line height with the threshold to eliminate big headings. Therefore, they yield higher accuracy due to the vertical rule line information.

TABLE I. EXPERIMENTAL RESULT OF SCANNED DOCUMENTS

| Category of Tables | Total Number of Scanned Documents | Correctly Detected Tables |
|---|---|---|
| Table with fully and partially bounding lines (Type A) | 110 | 91 |
| Table with parallel lines (Type B) | 135 | 91 |
| Table without ruling lines (Type C) | 53 | 40 |

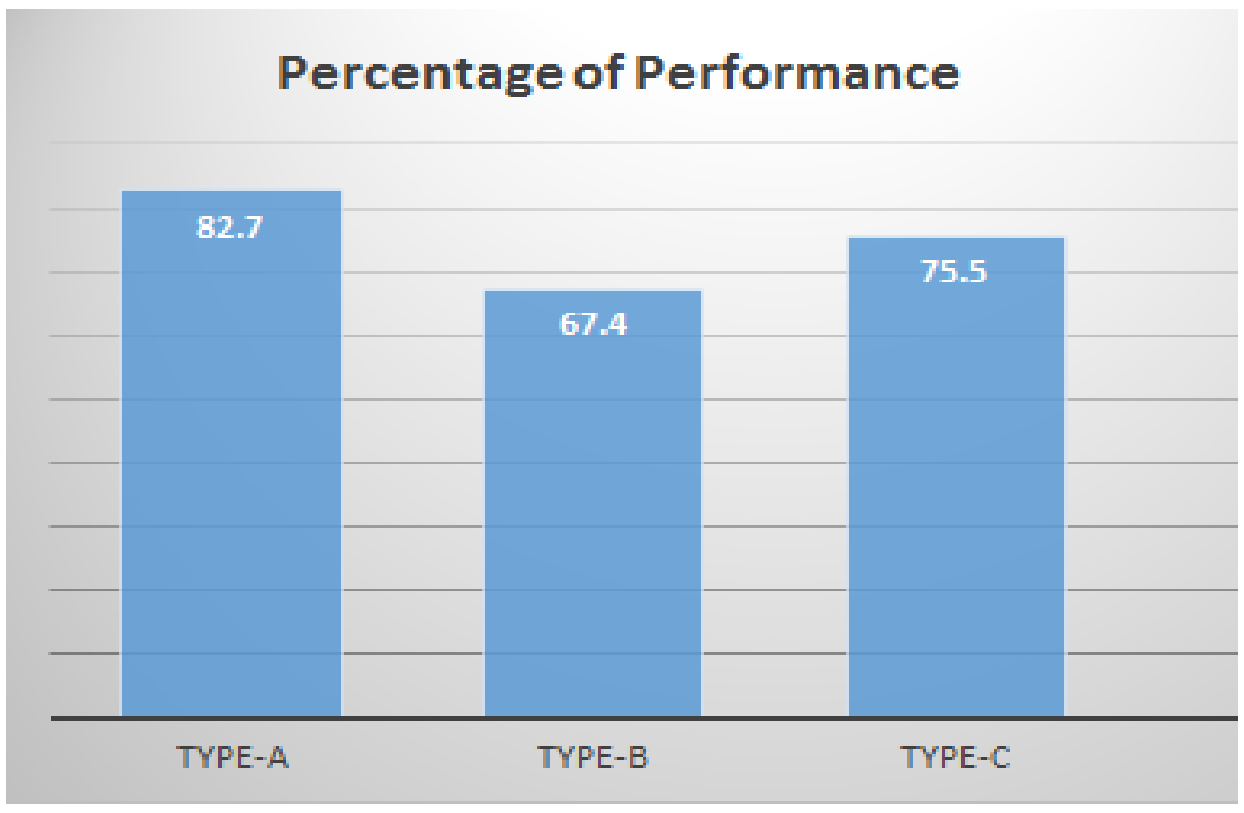

Fig. 8. Percentage of Performance of different categories of tables in document images

Among the collected documents, 53 have tables without any ruling lines (type C), in which 40 tables have been located. Seven documents cannot be located due to the differences with font type, size and style within the page and two-column table structure in which within a row one column have multiple text lines and the other column with single data. Another 6 documents cannot be identified due to the header and footer section in which section or document title and page number with large word space between them interfere the detection process.

Similarly, out of 135 tables with parallel lines, 91 documents are detected. The main reason for not to identify these tables is the assumption of the threshold for the word space made in the algorithm such that threshold is equal to 1.5 * word space. The second reason is header and footer section in which, section or document title and page number with space between them interferes the detection process in 5 documents.

Figures 9-14 depict the output produced by our algorithm for the three different types of input documents presented in Figures 2-7.

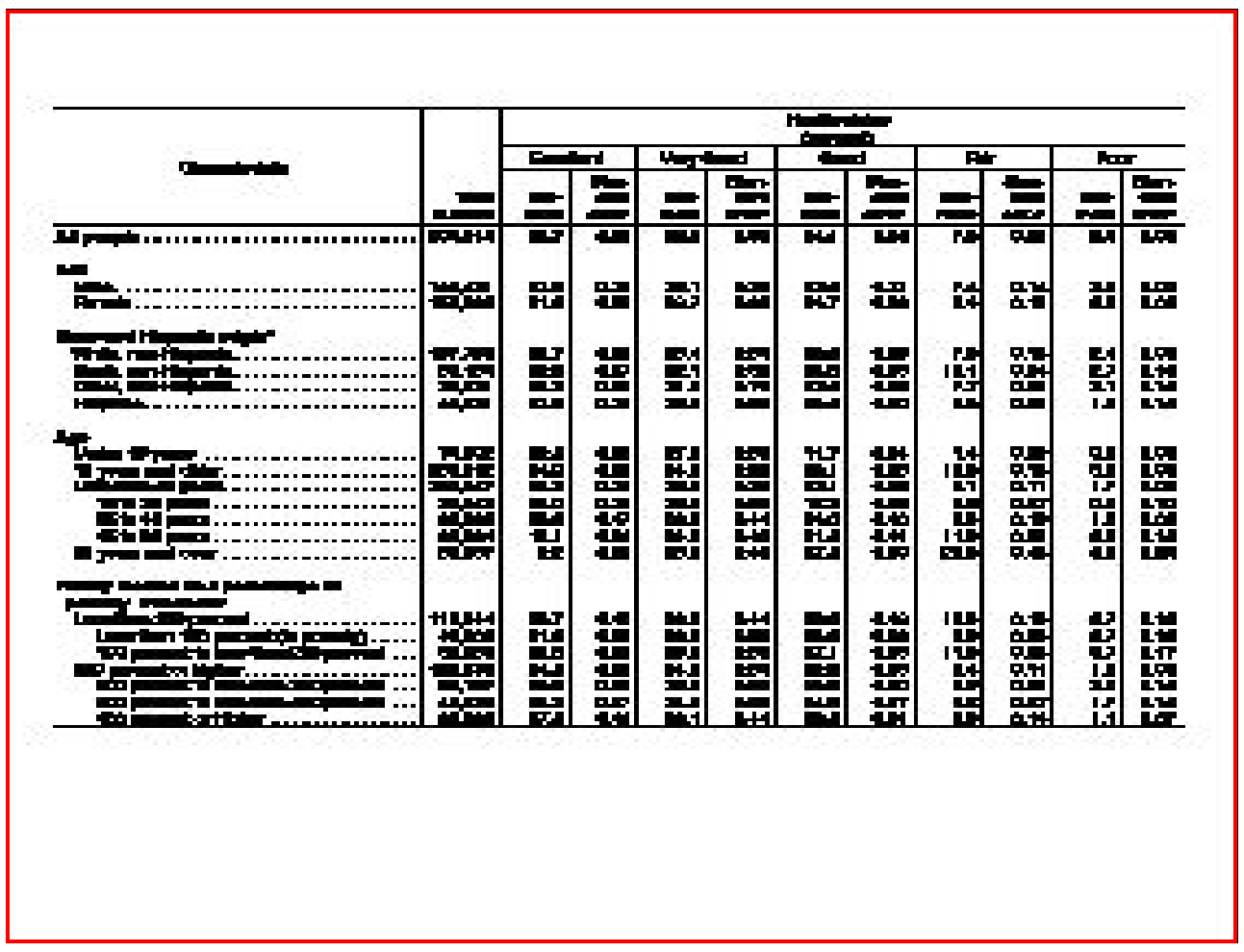

Fig. 9. Detected tables from document type A from Figure 2

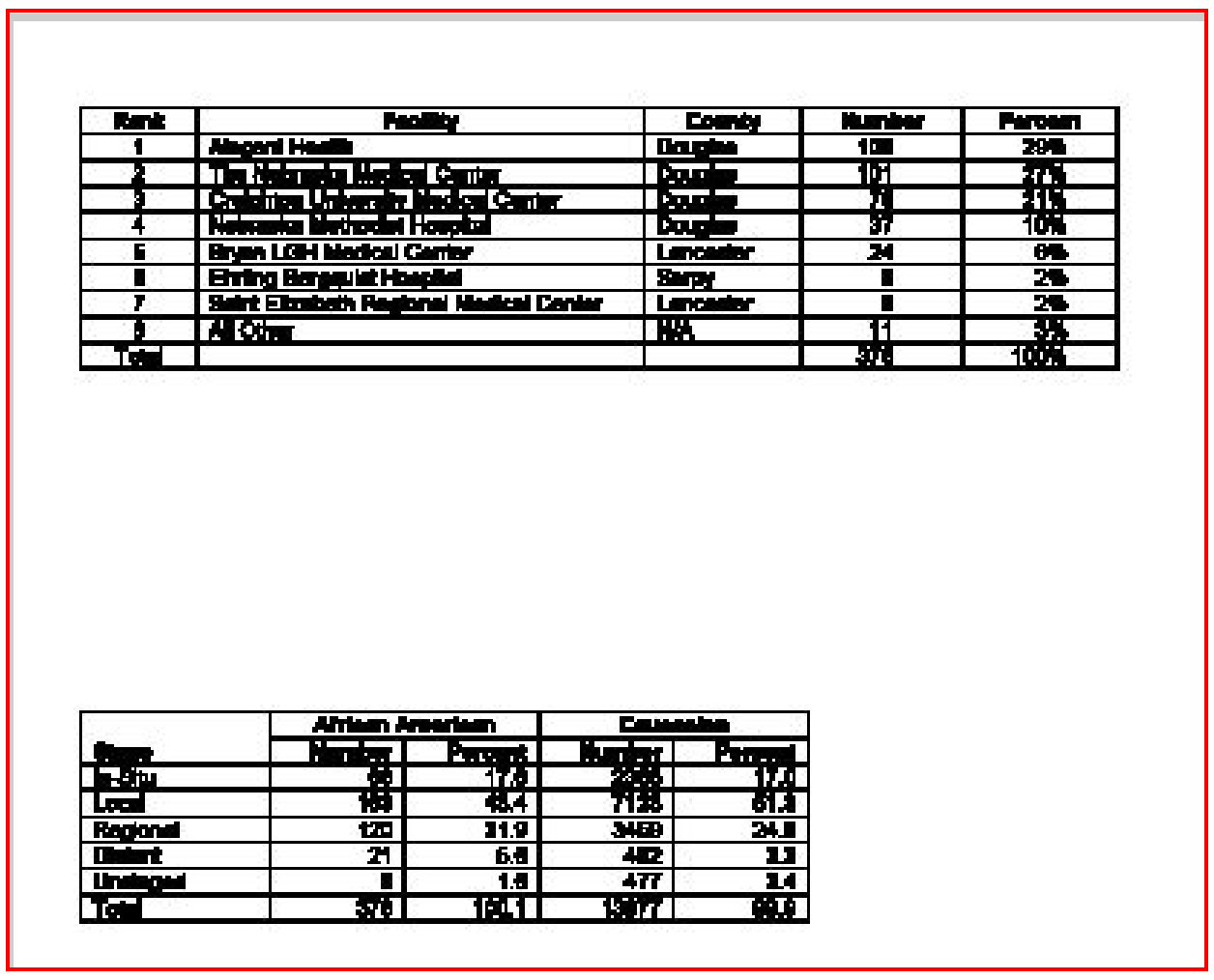

| Rank | Facility | County | Number | Percent |
|---|---|---|---|---|
| 1 | Alegent Health | Douglas | 108 | 29% |
| 2 | The Nebraska Medical Center | Douglas | 101 | 27% |
| 3 | Creighton University Medical Center | Douglas | 78 | 21% |
| 4 | Nebraska Methodist Hospital | Douglas | 37 | 10% |
| 5 | Bryan LGH Medical Center | Lancaster | 24 | 6% |
| 6 | Ehrling Bergquist Hospital | Sarpy | 8 | 2% |
| 7 | Saint Elizabeth Regional Medical Center | Lancaster | 8 | 2% |
| 8 | All Other | N/A | 11 | 3% |
| Total | | | 378 | 100% |

| | African American | | Caucasian | |
|---|---|---|---|---|
| Stage | Number | Percent | Number | Percent |
| In-Situ | 66 | 17.6 | 2388 | 17.0 |
| Local | 183 | 48.4 | 7128 | 51.3 |
| Regional | 120 | 31.9 | 3468 | 24.8 |
| Distant | 21 | 5.6 | 482 | 3.3 |
| Unstaged | 6 | 1.6 | 477 | 3.4 |
| Total | 378 | 100.1 | 13977 | 99.8 |

Fig. 10. Detected tables from document type A from Figure 3

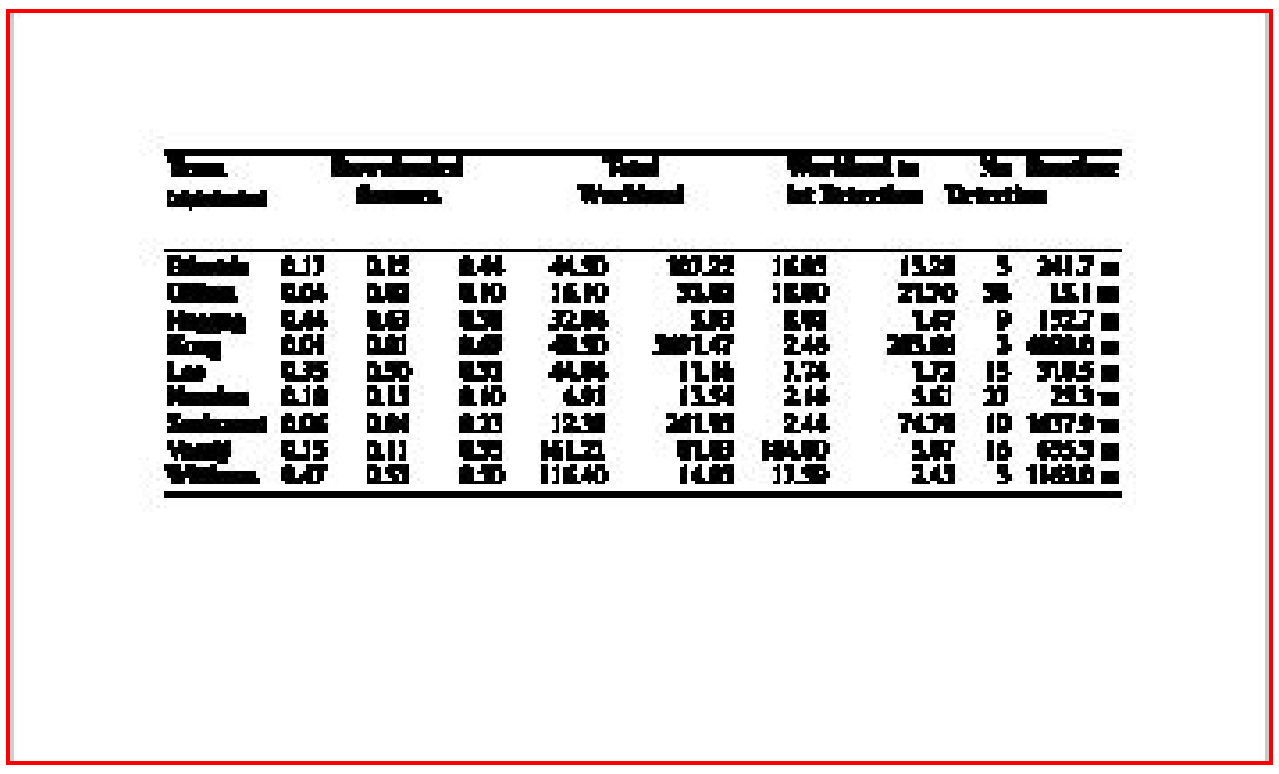

Fig. 11. Detected tables from document type B from Figure 4

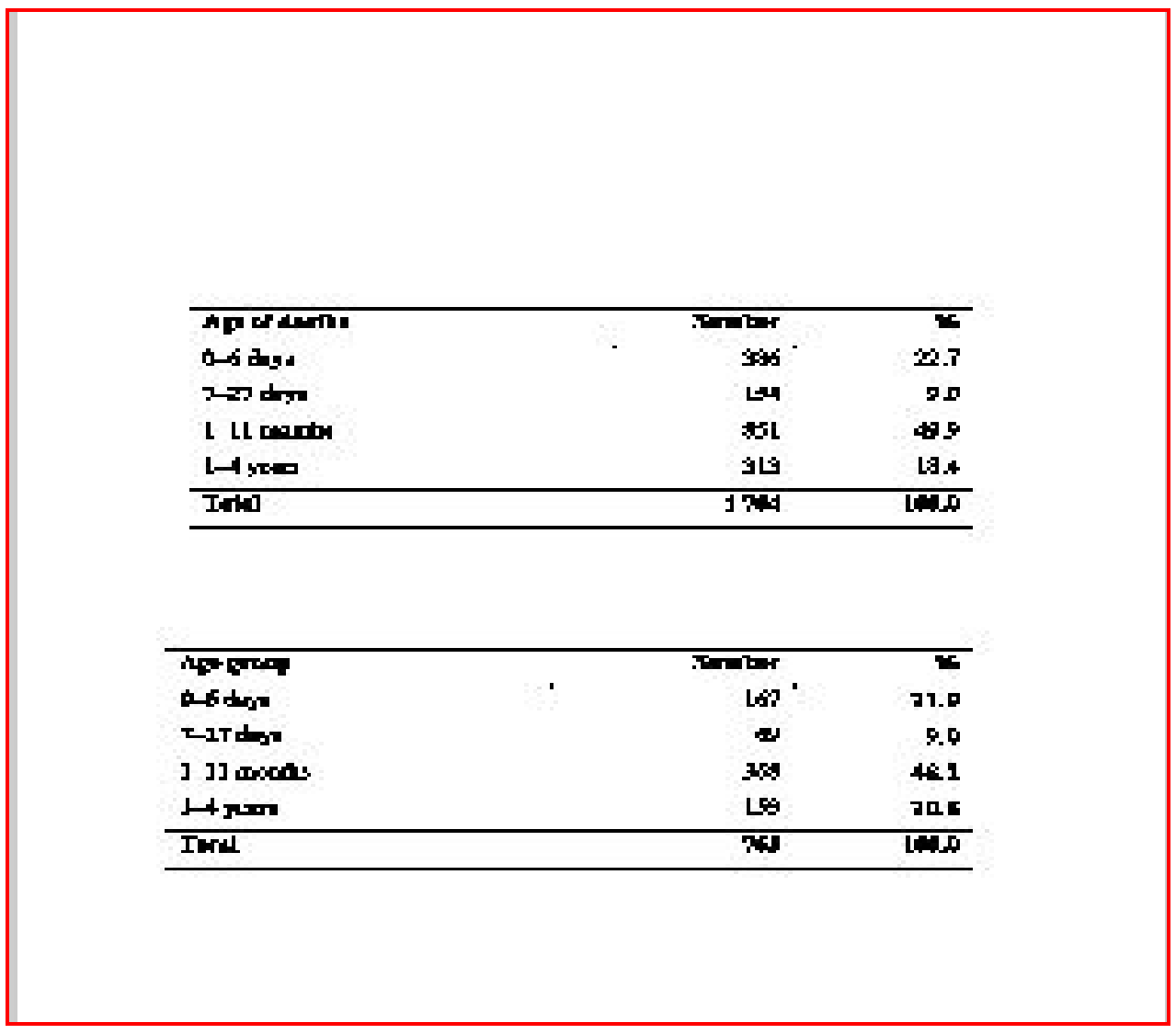

| Age of death | Number | % |
|---|---|---|
| 0–6 days | 396 | 22.7 |
| 7–27 days | 164 | 9.0 |
| 1 11 months | 851 | 48.9 |
| 1–4 years | 313 | 18.4 |
| Total | 1 744 | 100.0 |

| Age group | Number | % |
|---|---|---|
| 0–6 days | 167 | 21.9 |
| 7–27 days | 69 | 9.0 |
| 1 11 months | 358 | 46.1 |
| 1–4 years | 159 | 20.6 |
| Total | 763 | 100.0 |

Fig. 12. Detected tables from document type B from Figure 5

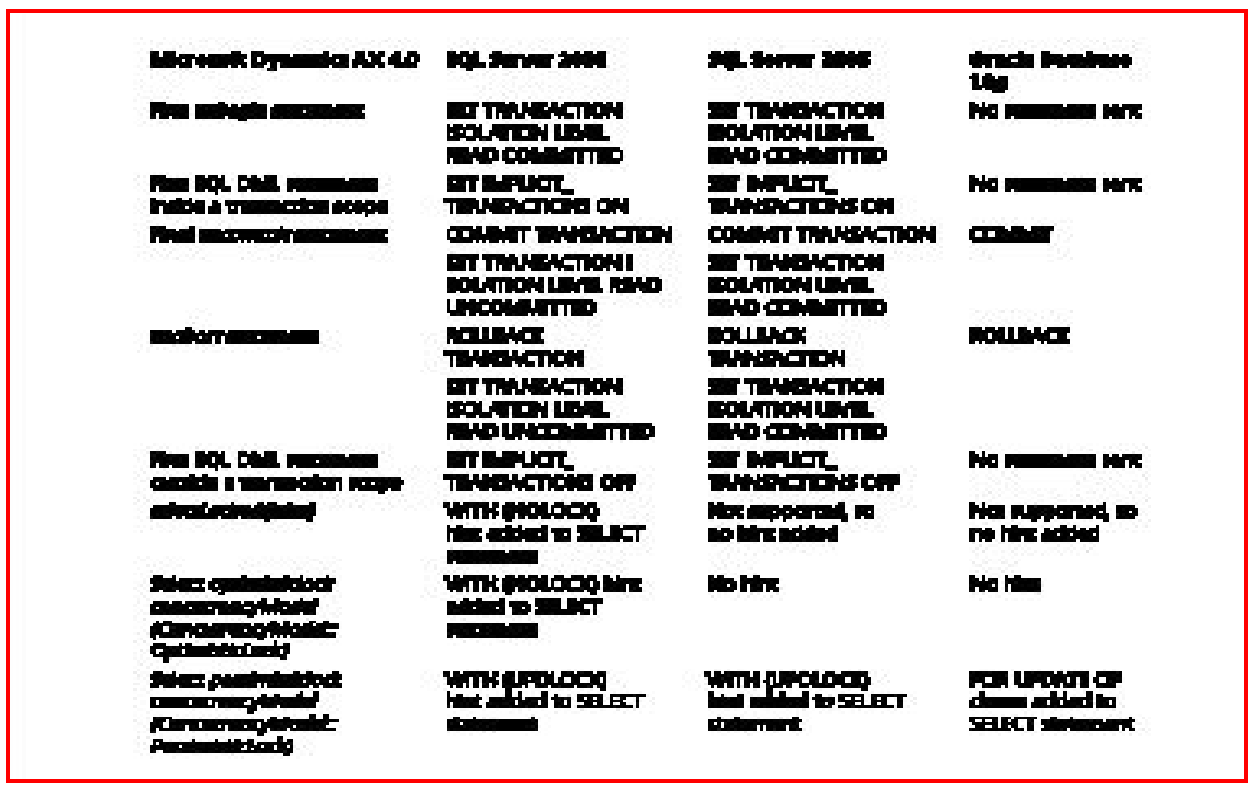

Fig. 13. Detected tables from document type C in Figure 6

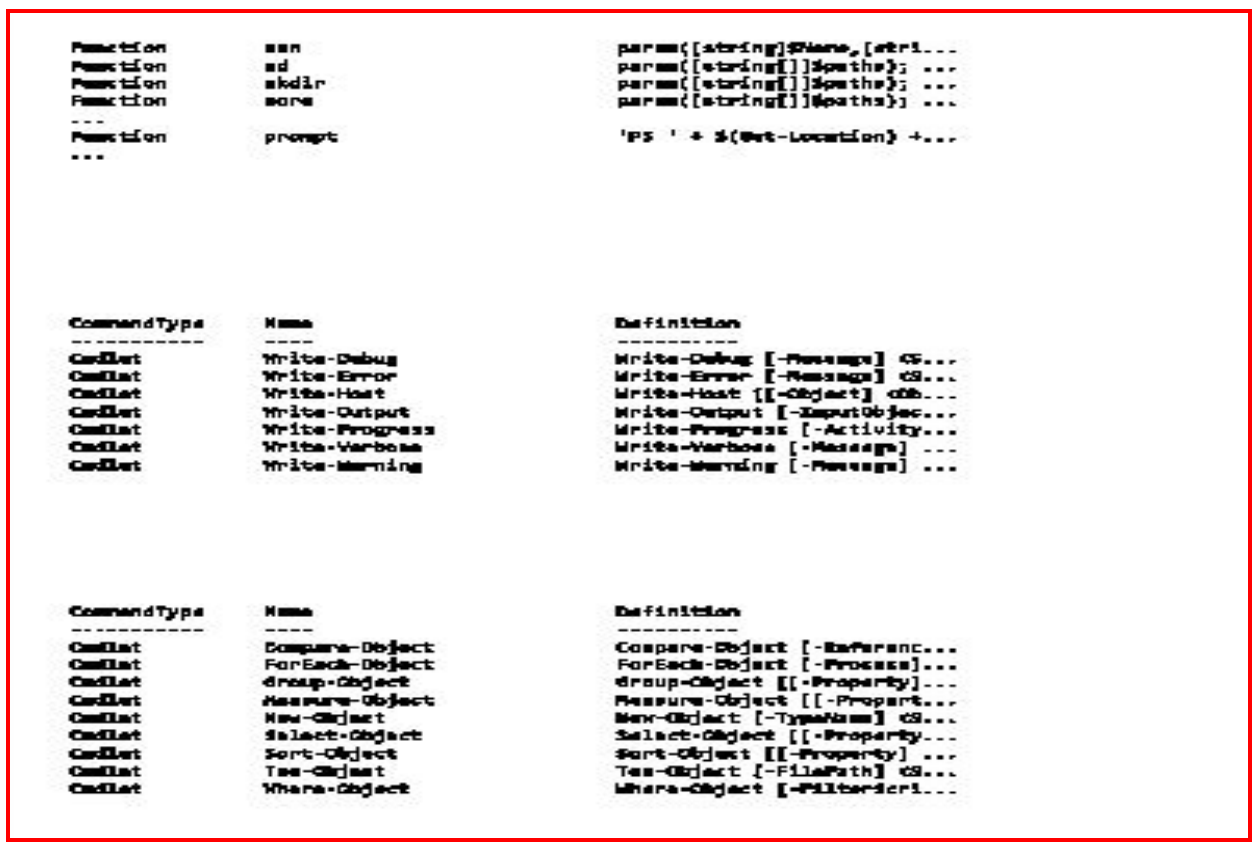

Fig. 14. Detected tables from document type C in Figure 7

## V. CONCLUSIONS

In this paper, we have proposed an algorithm for locating tables from scanned documents. From the experiments, the system has shown 75% overall detection rate for 298 documents in all different type of tables in ranges of scanned documents. The system produces erroneous results, when the document contains header or footer section or document title and page number in a line with large space between them. The algorithm can be applied to handle tables in multi-column layout with small modification in layout analysis. Since, we have determined a local threshold for line height and word space, which depends only on the particular document with its own format and style, this would be more advantageous in applying different layouts, formats in document images. Moreover, any layout of document with white and grey background also can be treated with this algorithm.

The future work of this research is to focus on (i) improving the accuracy of the system by eliminating the issues arise from headers and footers; and (ii) handling tables with decorated lines; (iii) extraction of features from the located tables to reconstruct and reproduce the documents with their existing formats.

## REFERENCES

[1] R. Zanibbi, D. Blostein and J.R. Cordy, "A survey of table recognition: models, observations, transformations, and inferences," International Journal of Document Analysis and Recognition, vol. 7, pp.1-16, 2003.

[2] T. Kieninger and A. Dengel, "Applying the T-RECS table recognition system to the business letter domain," In Proc. Int. Conf. on Document Analysis and Recognition, pp. 518-522, Seattle, WA, USA, 2001.

[3] Thomas G Kieninger, "Table structure recognition based on robust block segmentation".

[4] S. Mandal, S. Chowdhury, A. Das, and B. Chanda, "A simple and effective table detection system from document images," Int. Journal on Document Analysis and Recognition", vol.8(2-3), pp. 172-182, 2006.

[5] B. Yildiz K Kaiser and S. Miksch, "PDF to Table: A method to extract table information from pdf files," Proceeding of International conference on Artificial Intelligence (IICAI'05), 2005, pp 1773-1775.

[6] E. Oro, M. Ruffolo, "PDF-TREX: An approach for recognizing and extracting tables from PDF documents," Proceeding of International conference on Document Analysis and Recognition, pp. 906-910, 2009.

[7] Jing Fang, Liangcai Gao, Kun Bai, Ruiheng Qiu, Xin Tao, Zhi Tang, "A table detection method for multipage PDF documents via visual separators and tabular structures," Proceeding of International conference on Document Analysis and Recognition(ICDAR'11), 2011

[8] Y. Wang, R. Haralick, and I. T. Phillips, "Automatic table ground truth generation and a background-analysis-based table structure extraction method," In Proc. Int. Conf. on Document Analysis and Recognition, pp. 528-532, Seattle, WA, USA, 2001.

[9] T. Watanabe, H. Naruse, Q. Luo and S. Sugie, "Structure analysis of table-form documents on the basis of the recognition of the vertical and horizontal line segments," International conference on Document Analysis and Recognition, pp. 906-910, 2006.

[10] T. Laurentini, P. Vaida, "Identifying and understanding tabular material in compound documents," Proceeding of the international conference on pattern recognition, 1992.

[11] Edward A. Green, Mukki S. Krishnamoorthy, "Model-based analysis of printed tables".

[12] F. Cesarini, S. Marinai, L. Sarti, and G. Soda, "Trainable table location in document images," In Proc. Int. Conf. on Pattern Recognition, pp. 236-240, Canada, Aug. 2002.

[13] B. Gatos, D. Danatsas, I. Pratikakis, and S. J. Perantonis. "Automatic table detection in document images," In Proc. Int. Conf. on Advances in Pattern Recognition, pp. 612-621, Path, UK, Aug. 2005.

[14] A.C. e Silva, "Learning rich hidden markov models in document analysis: Table location," In Proc. Int. Conf. on Document Analysis and Recognition, pp. 843-847, Barcelona, Spain, July 2009.

[15] J. Hu, R. Kashi, D. Lopresti, and G. Wilfong, "Medium- independent table detection," In Document Recognition and Retrieval", VIII (IS&T/SPIE Electronic Imaging, pp. 44–55, 2001.

[16] F. Shafait and R. Smith, "Table detection in heterogeneous documents," In Proceedings of the 9th IAPR International Workshop on Document Analysis Systems, pp. 65–72, 2010.

[17] A. Namboodiri, "On-line handwritten document understanding," Ph.D thesis, Michican State University, 2004.

[18] J. Chen and D. Lopresti, "Table detection in noisy off- line handwritten documents," In Proceedings of the 11th International Conference on Document Analysis and Recognition, pp. 399–403, 2011.

[19] T. Kasar, P. Barlas, S. Adam, C. Chetalain and T, Pacquet, "Learning to detect tables in scanned document images using line information," 12th International Conference on Document Analysis and Recognition, 2013.

[20] S. Mandal, S.P. Chowdhury, A.K. Das, and Bhabatosh Chanda, "A complete system for detection and identification of tabular structures from document images," ICIAR 2004.

[21] Gaurav Harit, Anukritti Bansal, "Table detection in document images using header and trailer patterns," ICVGIP, December, 2012.

[22] Ray Smith, "Hybrid page layout analysis via tab-stop detection," 10th International conference on Document Analysis and Design (ICDAR), 2009.

[23] B .Gatos I.Pratikakis, S.J Perantonis, "An adaptive binarization technique for low quality historical documents," IARP Workshop on Document Analysis Systems (DAS 2004), pp. 102-113.